\pdfoutput=1

\documentclass[11pt]{article}

\usepackage{acl}

\usepackage{times}
\usepackage{latexsym}

\usepackage[T1]{fontenc}

\usepackage[utf8]{inputenc}

\usepackage{microtype}

\usepackage[subtle]{savetrees}
\usepackage{multirow}
\usepackage{hyperref}
\usepackage{booktabs} 
\usepackage{tabularx}
\usepackage{graphicx}
\usepackage{subcaption}

\usepackage[T1]{fontenc}
\usepackage{ragged2e}
\usepackage{siunitx}
\usepackage{adjustbox}
\newcommand*\rot{\rotatebox{90}}
\newcommand{\STAB}[1]{\begin{tabular}{@{}c@{}}\rot{#1}\end{tabular}}
\newcommand{\uid}{\textsf{\small UserIdentifier}}

\usepackage{titlesec}
\titlespacing*{\section}
{0pt}{1ex}{0.75ex}
 \titlespacing*{\subsection}
 {0pt}{0.5ex}{0.4ex}
 \titlespacing*{\subsubsection}
 {0pt}{.0ex}{.1ex}
 
\makeatletter
\renewcommand{\paragraph}{%
  \@startsection{paragraph}{4}%
  {\z@}{1.25ex \@plus .5ex \@minus .2ex}{-1em}%
  {\normalfont\normalsize\bfseries}%
}
\makeatother

\title{UserIdentifier: Implicit User Representations for Simple and Effective \\ Personalized  Sentiment Analysis \vspace{0ex}}

\author{Fatemehsadat Mireshghallah\textsuperscript{\rm 1}\thanks{\quad Work done as part of an MSR internship. Corresponding author email: fatemeh@ucsd.edu}, Vaishnavi Shrivastava\textsuperscript{\rm 2}, Milad Shokouhi\textsuperscript{\rm 2},\\
    \textbf{Taylor Berg-Kirkpatrick}\textsuperscript{\rm 1}, \textbf{Robert Sim}\textsuperscript{\rm 3}, \textbf{Dimitrios Dimitriadis}\textsuperscript{\rm 3}\\
    \textsuperscript{\rm 1} University of California San Diego,
    \textsuperscript{\rm 2} Microsoft Corporation, 
    \textsuperscript{\rm 3} Microsoft Research \\
    \texttt{[fatemeh, tberg]@ucsd.edu},\\ \texttt{ [vashri,milads,rsim,didimit]@microsoft.com}\\
    }

\begin{document}
\maketitle

\begin{abstract}
\vspace{-1ex}
Global models are typically trained to be as generalizable as possible. Invariance to the specific user is considered desirable since models are shared across multitudes of users. However, these models are often unable to produce personalized responses for individual users, based on their data.  Contrary to widely-used personalization techniques based on few-shot and meta-learning, we propose \uid, a novel scheme for training a single shared model for all users. Our approach produces personalized responses by prepending a fixed, user-specific non-trainable string (called ``user identifier'') to each user's input text. Unlike prior work, this method doesn't need any additional model parameters, any extra rounds of personal few-shot learning, or any change made to the vocabulary. We empirically study different types of user identifiers (numeric, alphanumeric, and also randomly generated) and demonstrate that, surprisingly, randomly generated user identifiers outperform the prefix-tuning based state-of-the-art approach by up to $13\%$, on a suite of sentiment analysis datasets. 
\end{abstract}


\section{Introduction}
\label{sec:intro}

\begin{figure}[h!]
    \centering
     \includegraphics[width=0.98\linewidth]{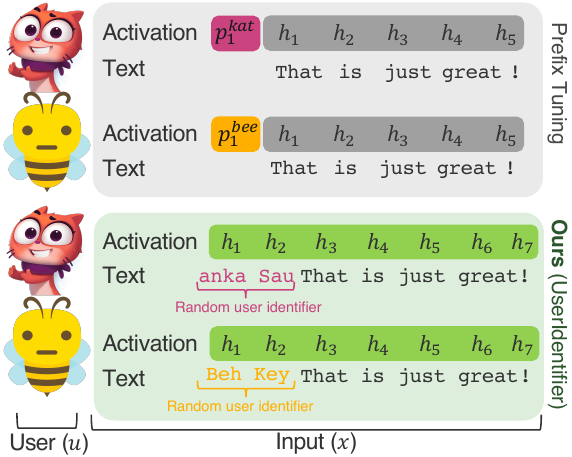}
     \caption{An overview of the proposed method, \uid, compared to its prefix-tuning counterpart. $p^{kat}_1$, $p^{bee}_1$ denote the trainable prefix vector for users $kat$ and $bee$, in the prefix tuning method~\cite{useradapter}. \uid, on the other hand, does not have trainable user-specific parameters and uses random per-user (UID) strings (``\texttt{anka Sau}'' and ``\texttt{Beh KY}''), to condition a shared model, for each user. }
     
     \label{fig:uid}
    \vspace{-3ex}
\end{figure}

%
Personalization arises in applications where different clients need models specifically customized to their environment and user profiles~\cite{yang-eisenstein-2017-overcoming,mazare-etal-2018-training,flek-2020-returning}.  
%
%
This need for customization stems from the inherent heterogeneity existing in the data and the labels, especially when the task is classification~\cite{kulkarni2020survey, wang-etal-2018-personalized}.
Fig.~\ref{fig:uid} shows an example of the sentence ``That is just great!''. This sentence could carry a positive sentiment, a neutral apathetic sentiment, or even a completely negative sentiment. A non-personalized model cannot correctly predict the label for different users.

Most techniques for personalization generally involve two phases: first, a shared, global model is built between all users, and then, it is personalized for each client using their data~\cite{kulkarni2020survey, Schneider2019MassPO,lee-etal-2021-meta}.
In such cases, each user has either an entirely separate model, or additional personal parameters, causing significant overheads, both in terms of storage of the large models, and the computation complexity of training separate models for each user.
UserAdapter~\cite{useradapter}, the state-of-the-art in personalized sentiment analysis, takes a prefix-tuning based approach~\cite{li-liang-2021-prefix}, as shown in Fig.~\ref{fig:uid}. In the first phase, a global model is trained in a user-agnostic way on a large dataset. 
In the second phase, each user $u$ is assigned their own prefix vector, $p_1^u$, which is fine-tuned separately for them, on their own data. If there are $N$ users, there would be $N$ separate rounds of fine-tuning, producing $N$ vectors. During this prefix-tuning phase, the underlying transformer-based classification model is frozen and shared between users, and the final $N$ vectors are stored for inference.

To alleviate these training and storage costs and also improve overall performance, we propose training a single, shared personalized model, which can capture  user-specific knowledge 
by conditioning on a unique, user-specific sequence of tokens from the classifier's vocabulary. We name this sequence ``user identifier'', and dub the underlying method of adding user identifiers to the input \uid{}. This is shown in Fig.~\ref{fig:uid}, where we add the randomly generated, and non-trainable user identifiers ``\texttt{anka Sau}'' and ``\texttt{Beh KY}'' to each user's sample, and then train the transformer classifier model, on these augmented samples. 
%
The user identifiers just use the underlying model's vocabulary and embeddings and do not add any tokens nor any user embeddings to the model. They are also static over time, and unique to each user, which means the user ``bee'' in Fig.~\ref{fig:uid} will have ``\texttt{Beh KY}'' pre-pended to all their samples, and no other user has this identifier.
This is similar to the prompting of models like GPT-3~\cite{brown2020language}, however, here the prompt is fixed and used as data augmentation during training, and the model is not generative.
As such, we only do training once and have one set of shared parameters for all users. ~\textcolor{black}{The approach is similar in essence to those  of~\citet{daume2009frustratingly,KOCON2021102643,kocon2021learning},
which augments each individual feature with domain annotations. }

We experiment with different types of strings for user identifiers, such as real usernames from the dataset, consecutive numbers, random digits, random non-alphanumeric tokens, and random tokens (all types), and observe that, surprisingly, random identifiers, sampled from all possible tokens in the vocabulary perform best, providing $1.5\%-13\%$ classification accuracy improvement on average, over the prefix-tuning based method UserAdapter~\cite{useradapter}.
We also study different lengths of identifiers. We report our results on three different sentiment analysis datasets (Sentiment 140, IMDB, and Yelp). 
We also show that~\uid{} is effective in a federated learning setup (Appendix~\ref{sec:fl}), which is a real-world application of such personalization~\cite{kulkarni2020survey}. 

%
%
%

\section{UserIdentifier}

In this section, we  first explain how \uid{} operates, then we go over the parameterization and learning procedure. 

\subsection{Method}
\uid{} is a data augmentation method which consists of adding a sequence of user-specific tokens (user identifier, $u_{id}$, drawn from the tokenizer's vocabulary) to each  sample, $x$, to provide  user-related cues to the model and help it learn individual user behaviour and preferences, all in one shared model. 
Figure~\ref{fig:uid} shows how this augmentation works. Each utterance is appended by the user identifier to create the augmented sample $[u_{id};x]$, and then  used as input to the model, for the training stage. 

There is no restriction on what the make-up or the length of the user identifier sequence can be (as long as it is not longer than the maximum sequence length the model can input). However, we propose randomly generating each user's  identifying sequence, through uniformly sampling from the tokenizer  vocabulary,  for a given length $L$, which we ablate in section~\ref{sec:abl}. This random sampling step creates a diverse while unique set of user identifiers, potentially allowing the model to distinguish different users more efficiently. 

\subsection{Parameterization}

For parameterizations of the user identifiers, we use parameter tying~\cite{he2019probabilistic}, where the user identifiers use the same set of parameters for their embeddings as the rest of the user utterance. In other words, in this setup the user embedding parameters are tied to the  embedding parameters of the main transformer classification model, parameterized by $\theta$. This form of parameterization is both simpler and has highere performance (we try separate parametrization in our experiments and show its inferior performance).

\subsection{Learning}
The training  stage doesn't change compared to the original fine-tuning process, with  parameters $\theta$ of the transformer model being trained to minimize the cross-entropy loss for the classification~\cite{devlin2018bert}:

\begin{equation}
 \mathcal{L}_{\textsc{CE}}(x,u_{id},y;\theta)=  - \log \Pr(y | [u_{id};x] ; \theta)   
\end{equation}
\begin{equation}
 \theta = \mathop{\arg \min}\limits_{\theta} \;\mathcal{L}_{\textsc{CE}}(x,u,y;\theta)  
\end{equation}

Where $x$ denotes the input utterance, $u_id$ denotes the user identifier for the user to whom utterance $x$ belongs, and $y$ is the class label for  $x$.

\section{Experimental Setup}
\begin{table}[]
    \centering
    \caption{Dataset specifications}
      \vspace{-2ex}
    \label{tab:data}
    \begin{adjustbox}{width=\linewidth, center}
     \newcolumntype{L}{>{\RaggedLeft\arraybackslash}p{0.06\linewidth}} 
  \newcolumntype{O}{>{\RaggedLeft\arraybackslash}m{0.07\linewidth}} 
  \newcolumntype{D}{>{\arraybackslash}m{0.15\linewidth}} 
  \newcolumntype{R}{>{\arraybackslash}m{0.29\linewidth}} 
   \newcolumntype{K}{>{\raggedleft\arraybackslash}p{0.22\linewidth}} 
\begin{tabular}{@{}llKKK}
	\toprule
	& {{Dataset}} & {{\# Users}} &{{\# Samples}} &{{\# Classes}} \\

    \midrule
          & IMDB	&1,012	&137,710	& 10	 \\
    &Yelp	        &4,460	&428,369	& 5	\\
    &Sent140	        &1,100	& 56,557		&2 	\\
    &Sent140 (skewed)	        &473	& 23,155	&2		\\
	\bottomrule
\end{tabular}
    \end{adjustbox}
        \vspace{-2ex}
\end{table}

\begin{table*}[t]
    \centering
    \caption{Comparison of sentiment classification accuracy of \uid{}, with the baselines of Section~\ref{sec:baselines}.  Num., Def. and Rand. refer to the different types of user identifiers introduced in Section~\ref{sec:type}. }
    \vspace{-1ex}
    \label{tab:cent}
    \begin{adjustbox}{width=\textwidth, center}
     \newcolumntype{L}{>{\RaggedLeft\arraybackslash}p{0.06\linewidth}} 
  \newcolumntype{O}{>{\RaggedLeft\arraybackslash}m{0.07\linewidth}} 
  \newcolumntype{D}{>{\arraybackslash}m{0.15\linewidth}} 
  \newcolumntype{R}{>{\arraybackslash}m{0.29\linewidth}} 
\begin{tabular}{@{}clSSSSScSSSS@{}}
	\toprule
	& {\multirow{2}{*}{Dataset}} & {\multirow{2}{*}{Conventional}} &{\multirow{2}{*}{UserAdapter}}	& \multicolumn{3}{c}{{Trainable User Emb.}}	 & \text{} & \multicolumn{3}{c}{{\uid}}	 \\
	\cmidrule{5-7} \cmidrule{9-11}  
	&                       &                            &                          & {Num.} & {Def.} & {Rand. All} && {Num.} & {Def.}  &  {Rand. All}  & \\
    \midrule
    \multirow{2}{*}{\STAB{ \scriptsize  RoBERTa}} & 
    IMDB	&45.1 &	46.2&  		45.5&{--}&	48.9&		& 50.1 &{--}&	\textbf{52.5} \\
    &Yelp&	68.3	&70.2	& 68.3&{--}	&70.6	&&69.5&	{--}& \textbf{71.3}	\\
    \midrule[0.1pt]
    \multirow{2}{*}{\STAB{\scriptsize BERT}} &
    Sent140	&84.7&  {--}	&	84.7&	86.3&	86.5&&	84.9&	87.1&	\textbf{87.1}	\\
    &Sent140 (Skewed)	 &       86.3& {--} &	87.2&	89.3	&90.0	&&87.5	&90.3	&\textbf{90.4}	\\

	\bottomrule
\end{tabular}

    \end{adjustbox}
        \vspace{-2ex}
\end{table*}
\begin{table}[t]
    \centering
    \caption{Classification accuracy vs the length (\#tokens) and type (Section~\ref{sec:type}) of user identifier sequence) }
    \vspace{-2ex}
    \label{tab:ablate}
    \begin{adjustbox}{width=\linewidth, center}
     \newcolumntype{L}{>{\RaggedLeft\arraybackslash}p{0.06\linewidth}} 
  \newcolumntype{O}{>{\RaggedLeft\arraybackslash}m{0.07\linewidth}} 
  \newcolumntype{D}{>{\arraybackslash}m{0.15\linewidth}} 
  \newcolumntype{R}{>{\arraybackslash}m{0.29\linewidth}} 
\begin{tabular}{@{}clcSS@{}}
	\toprule
	& {{Seq. Len.}} & {{Rand. Dig}} &{{Rand. Non.}}	& {{Rand. All}} \\

    \midrule
    \multirow{5}{*}{\STAB{IMDB}} & 
    5    & 48.8	&51.3	&52.2		 \\
    &10	        &47.4	&51.7	&52.5		\\
    &20	        &47.1	&50.2	&51.1		\\
    &50	        &46.5	&48.7	& 50.8		\\
    &200	        &33.3	&32.8	& 40.1		\\
    \midrule[0.1pt]
    \multirow{5}{*}{\STAB{Yelp}} & 
    5    & 68.6	&69.3	& 70.8		 \\
    &10	        &68.7	&69.6	&71.3		\\
    &20	        &68.4	&68.6	&71.0		\\
    &50	        &67.8	&69.0	&70.6		\\
    &200	        &63.2	&60.2	& 65.1		\\

	\bottomrule
\end{tabular}
    \end{adjustbox}
    \vspace{-2ex}
\end{table}

\subsection{Tasks, Datasets, and Models}
We evaluate the proposed method on the task of sentiment analysis. Table~\ref{tab:data} shows a summary of the datasets used in our experiments. We use the IMDB~\cite{imdb} and Yelp~\cite{yelp} datasets for comparison with the  UserAdapter method~\cite{useradapter} and  for the ablation studies. 
Each user's data is split into train, test, and validation sets, with $0.8$, $0.1$, and $0.1$ ratios. 
For comparison purposes, we are using a subset of the available users, i.e. those with fewer than $50$ samples, as done by~\citeauthor{useradapter} in support of few-shot learning, for reporting test accuracy.
%
%
We use the RoBERTa-base model for this set of experiments.

In addition to IMDB and Yelp, we also report the performance of the proposed method on the Sentiment140 dataset~\
cite{sent140, caldas2018leaf}, which is a set of Tweets collected from Twitter and labeled positive or negative based on the emojis in each Tweet. 
For this dataset, We use the methodology provided by~\citet{fairfl} to preprocess and partition this dataset. 
We create a second version of this dataset, and mark it as ``skewed''. For this skewed data, the users have been selected such that their sentiments are mostly skewed, i.e. we only include users with $80\%$ or more positive or negative Tweets.  We do this to create a setup where data is more heterogeneously distributed. We use BERT-base-uncased for evaluations on the Sentiment140 dataset. 


\subsection{Baselines}\label{sec:baselines}

\paragraph{Conventional Training.} Conventional finetuning of the pre-trained transformer model on the full dataset, without personalization.

\paragraph{UserAdapter.}   In UserAdapter, the work closest to ours,  a per-user embedding is learned through few-shot learning and stored. These personal vectors are prepended to the users' data to create personal responses. This work proposes prefix-tuning~\cite{li-liang-2021-prefix} on a user-level. Unlike our method, UserAdapter consists of two phases, as discussed in the introduction.
%
%
\paragraph{Trainable User Embeddings.} \uid{} uses the same set of parameters (BERT embeddings) for embedding both the sample content and the user identifiers.  In other words, the text and user embedding parameters are tied. To untie these parameters, we introduce a third baseline, with trainable user embeddings. In this setup, while the tokens used for the user identifier are still drawn from the pre-trained model's tokenizer vocabulary, we're creating and training a separate set of global parameters for the user embedding, instead of using the pre-trained model's embedding. \textcolor{black}{These extra embedding parameters are placed in parallel to the model's existing embedding layer. Each input sequence is partitioned to the content and the UID, the content is fed to the model's existing embedding layer and the UID is fed to the new embedding.} 

\subsection{Types of User Identifiers} \label{sec:type}
 We investigate five scenarios (types of sequences) for the user identifiers. The length of the user identifier sequences can vary in terms of the number of tokens ($L$) for the last three of these scenarios. 
%

\noindent\textbf{Default (Def.)}: This scenario uses the real user id (e.g., username) of that user, when provided by the dataset and if they are not private. We only have this option available for the Sentiment140 dataset. 

\noindent\textbf{Consecutive Numbers (Num.)}: We assign each user a unique  number, from $1$ to $N$, representing each user (up to $N$ users).

\noindent\textbf{Random sequence of digits (Rand. Dig.)}: In this scenario,  $L$ independent and identically distributed (i.i.d) samples from the set of digits ($0$ to $9$) are drawn, creating a sequence of length $L$ for each user.

\noindent\textbf{Random sequence of tokens with non-alphanumeric characters (Rand. Non.)}:  $L$ i.i.d samples are drawn from a subset of tokens (with size $400$)  that contain non-alphanumeric characters, e.g., the token ~\texttt{Ã""}. The motivation for this scenario is that such user identifiers might be easier for the model to distinguish from the text (if we make sure the textual content in the sample has no overlapping tokens with the identifier).

\noindent\textbf{Random sequence of all tokens (Rand. All)}: This scenario draws  $L$ i.i.d samples from the set of all available tokens in the tokenizer vocabulary.

\vspace{-0.6ex}
\section{Results}
\vspace{-0.7ex}
 
 Apart from the evaluations here, We have also provided evaluations of applying our method to federated learning in Appendix~\ref{sec:fl}, and applying it to new unseen user samples in~\ref{sec:unseen}.
%

\subsection{Comparison with Baselines}
 A comparison of \uid{} with the state-of-the-art UserAdapter method, and the other baselines  is presented in Table~\ref{tab:cent}. 
For the \textbf{Num.} (consecutive numbers) and \textbf{Def.} (default username) scenarios, as detailed in Section~\ref{sec:abl}, the length of the user identifier sequences depends solely on the tokenization process. For the case of \textbf{Rand. All} (randomly sampled from all vocabulary tokens), however,  it is shown that the sequence length of $10$ tokens provides the best performance through the ablation study, therefore the results are reported for this length. Since the default usernames for IMDB and Yelp datasets are not provided, the corresponding results are not reported here.

 It is shown that \uid{} with randomly generated identifiers outperforms all baselines, in all tasks. Our intuition is that \uid{} outperforms UserAdapter because of collaborative learning and personalization happening simultaneously, unlike in the case of UserAdapter where personalization is performed separately for each user. 
The performance of trainable user embeddings appears inferior to that of \uid{}, which could be attributed to the parameter tying used in \uid{}. This parameter tying couples the learning problems for both domains (user identifier and text) and allows us to jointly learn from the full data, as in~\cite{he2019probabilistic}.
For the Sentiment140 dataset, we can see that increasing the heterogeneity or skew in the dataset boosts the benefits brought about by \uid{}.  This shows that the proposed method performs better in setups where personalization is actually needed~\cite{deng2020adaptive}. 

\subsection{Ablation Studies}\label{sec:abl}
%
Table~\ref{tab:ablate} shows our ablation study into the length and the type of the user identifier sequence, for IMDB and Yelp datasets. 
 The most evident trend is that performance significantly degrades in both datasets when the length of the user identifier sequence exceeds $20$ tokens, holding for all identifier types. This is because the length of the input text itself is essentially decreased (the maximum sequence length for RoBERTa is $512$, and the textual content of the sample is truncated to fit the user identifier) when increasing the length of the identifier. This decreases the useful information which could be used to infer sentiment, and in turn, it has an adverse effect on accuracy. 

 A rather surprising observation is that randomly sampling from the tokenizer's entire vocabulary outperforms sampling only from digits or from the non-alphanumeric tokens. 
This can be attributed to the different sizes of the sampling spaces for these three types, and the probability of overlap in user identifier from user to user.
%
%
For the random digits (\textbf{Rand. Dig.}) the sample space size for each token position is $10$, the number of possible digits. For the non-alphanumeric tokens, we have limited them to $400$, and for the token type all (\textbf{Rand. All}), the possible sample space is $47,400$. This means that the probability of having token overlaps in user identifiers is much much smaller in the last scheme than it is for the other two, or in other words, the hamming distance between different user identifiers is higher with this method.

One hypothesis that might explain the success of random user identifiers: random user identifiers are similar to random feature projections \cite{rahimi2007random}, but, in contrast with learnable embeddings, they are defined in terms of the pre-trained model's original token embeddings. This may have a positive effect on optimization during fine-tuning. 

\subsection{\textcolor{black}{User-level Study Accuracy}}
\begin{figure}[!htb]
     \centering
     \begin{subfigure}[h]{0.33\textwidth}
         \centering
         \includegraphics[width=\textwidth]{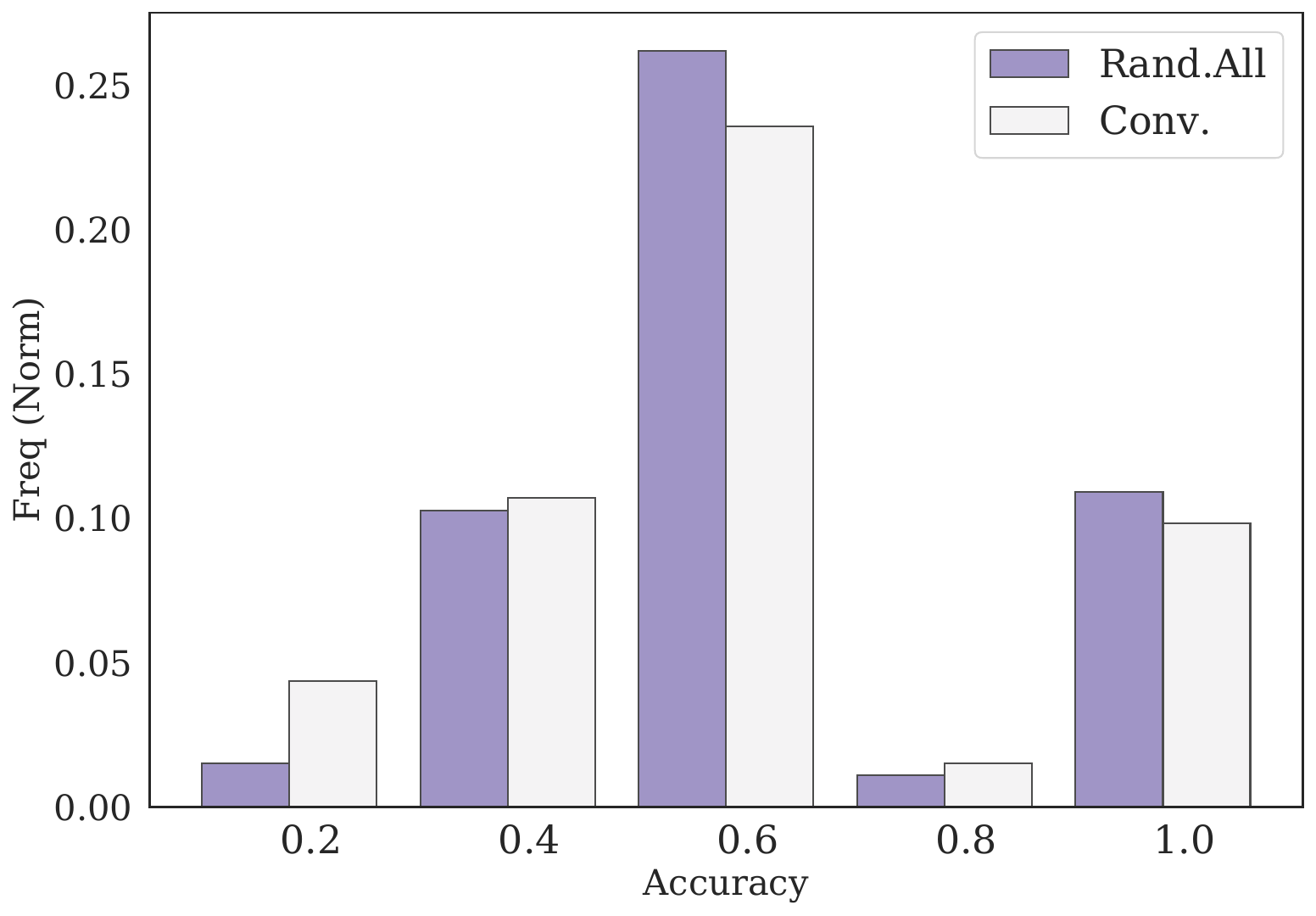}
         \caption{IMDB}
         \label{fig:dist:imdb}
     \end{subfigure}
     
     \begin{subfigure}[h]{0.33\textwidth}
         \centering
     \includegraphics[width=\textwidth]{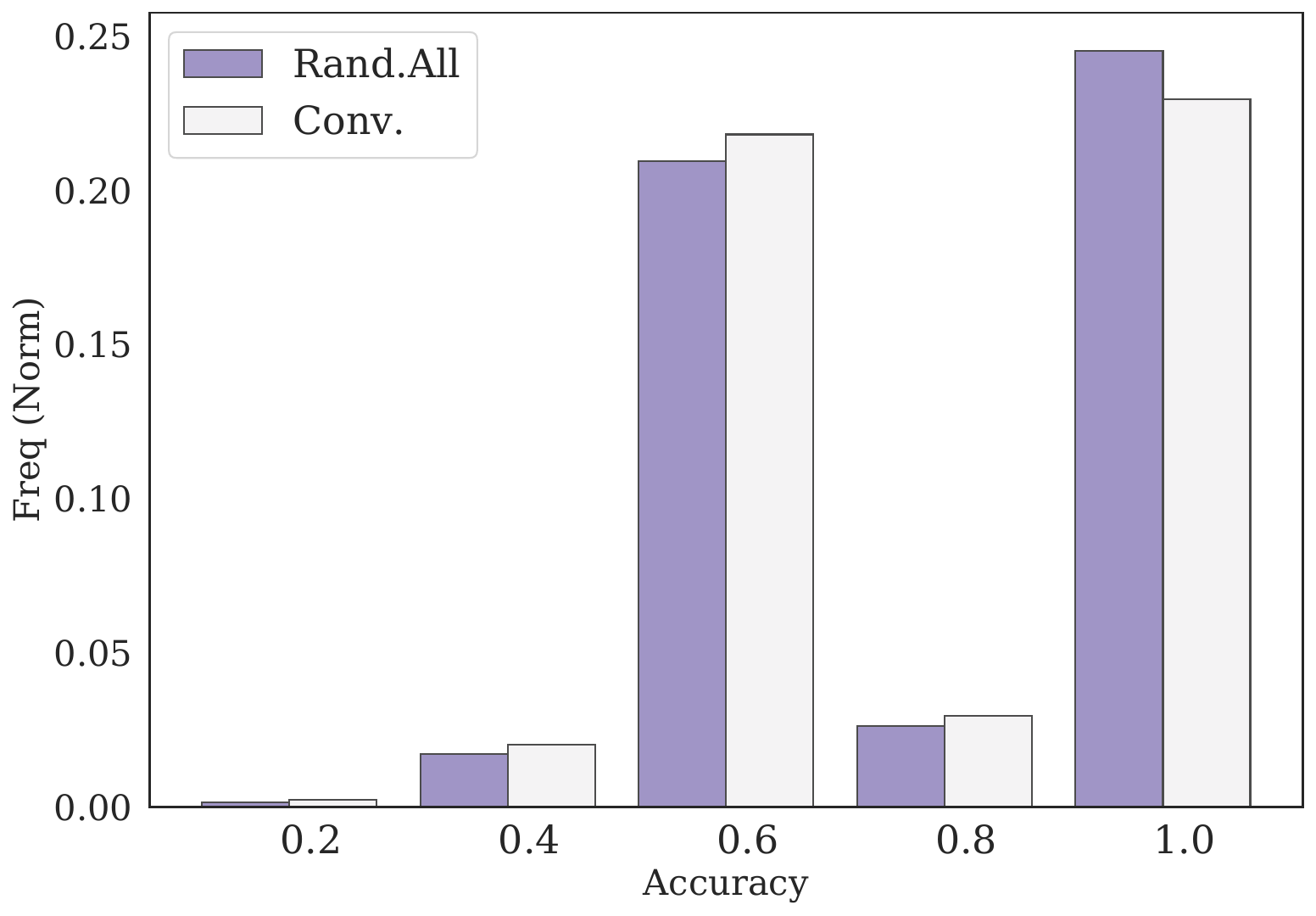}
         \caption{Yelp}
         \label{fig:dist:yelp}
     \end{subfigure}
     \vspace{-1ex}
     \caption{Distribution of test accuracy across users.
    }
    \vspace{-2ex}
     \label{fig:dist}
\end{figure}

\textcolor{black}{
Figure~\ref{fig:dist} shows the distribution of test-accuracy changes across users, for conventional training (Conv.) and the Rand.\ All scheme from \uid{}. We have chosen the best version of our model from Table~\ref{tab:cent} for this figure. 
We can see that the number of users with low accuracy decreases in both datasets. 
Also, the standard deviation of accuracy across users decreases compared to conventional training when using \uid{}, it drops from $27.0\%$ to $25.6\%$ for IMDB, and from $21.2\%$ to $21.0\%$ for Yelp. We provide more plots and analysis on this in~\ref{sec:change}.}

%

\subsection{\textcolor{black}{Performance on Unseen Users}}\label{sec:unseen}

To measure how robust the proposed method is to new users that have never been seen before, we run an evaluation on new users and report the results in Table~\ref{tab:unssen}. For this experiment, we have used the best models from Tables~\ref{tab:cent}, and tested them on samples from new users, without appending any user identifiers. It is noteworthy that there is some distribution shift between these unseen users and the seen users from Table~\ref{tab:cent}, especially for Yelp, as we used samples that were not used in the original training/test/val setup (this test set contains 5000 samples for Yelp and 1357 samples for IMDB). 

The \uid{} column refers to the accuracy of those datapoints on models trained with user identifiers, and the conventional column shows the accuracy but on a conventionally trained model, which would be the baseline. We can see that both models behave similarly, which suggests that for unseen data points, the \uid{} trained model falls back to a conventional model, and does not behave even worse.

\begin{table}[t]
    \centering
  \footnotesize
  \fontsize{7}{7}
   \renewcommand{\arraystretch}{0.6}
    \caption{Evaluation results on unseen users.}
   \vspace{-2ex}
    \label{tab:unssen}
    \begin{adjustbox}{width=\linewidth, center}
    \newcolumntype{L}{>{\RaggedLeft\arraybackslash}p{0.06\linewidth}} 
  \newcolumntype{O}{>{\RaggedLeft\arraybackslash}m{0.07\linewidth}} 
  \newcolumntype{D}{>{\arraybackslash}m{0.15\linewidth}} 
  \newcolumntype{R}{>{\arraybackslash}m{0.29\linewidth}} 
\begin{tabular}{@{}lccSS@{}}
	\toprule
	& {{\uid Accuracy (\%)}}	& {{Conventional Model Accuracy (\%)}} \\

    \midrule
    
 IMDB  & 50.4  	&50.9		 \\
    \midrule[0.1pt]
 
  Yelp &  50.1   	& 49.8		 \\

	\bottomrule
\end{tabular}
  \end{adjustbox}
\end{table}






\section{Conclusion}

 In this work, we present a novel approach for learning global models,  producing personalized classification responses. This method which doesn't require  model extensions or specialized training algorithms,
consists of appending a fixed, non-trainable, unique identifier string to each sample during training and inference. 
%
%

\section*{Acknowledgments}
The authors would like to thank the anonymous reviewers and meta-reviewers for their helpful feedback. We also thank Huseyin Inan and Guoqing Zheng for insightful discussions and Wanjun Zhong for helping with datasets. Additionally, we thank our colleagues at the UCSD and Microsoft for their helpful comments and feedback.

\section*{Ethical Considerations}
Our  proposed  model  is  intended  to  be  used  for addressing the problem of personalization, by learning one shared model for all users, and querying it using a personal identifier. One potential measure that needs to be taken for deployment of such technology is to setup proper authentication tools, so that each user can  only query with their own identifier  and prevent users from breaching privacy by querying other users' models. However, this could be a concern in other personalization setups too. 

The datasets used in our experiments are all publicly available (Yelp, IMDB and Sentiment 140), and we have not collected any information about the users who have contributed their data beyond what is originally provided in the dataset, which is only the user-based partitioning of the data.

\bibliography{anthology,custom}
\bibliographystyle{acl_natbib}


\appendix
\clearpage
\section{Appendix}

\subsection{Federated Learning as an Application}
\label{sec:fl}
Federated learning is a form of distributed learning where data never leaves each user's device~\cite{wang2021field,konevcny2018federated,Mireshghallah2020PrivacyID,basu2021benchmarking}. Instead, the user trains a model on their device locally and then shares the gradients (model updates) with a centralized server, which aggregates the gradients from different users and sends the updated model back to all of them, for further training. 
We target this setup since it is a good candidate for personalization, given how a conventionally trained global model often fails to accommodate all users~\cite{kulkarni2020survey,mansour2020three}. 
Table~\ref{tab:fl} shows the performance gain of applying \uid{}, in a federated setup.
\uid{} can be readily applied in federated learning, by assigning identifiers to each user and then asking them to append it to all their samples.  We have used the Rand.\ All type of user identifier for this experiment, since we observed in previous sections that it was the most effective. 
In general, the baseline performance and the performance gain in the federated setup is slightly lower than in centralized learning, which is due to the distributed nature of FL, and the fact that only the average of multiple gradient updates are shared with the server for aggregation.

\begin{table}[htb!]
    \centering
    \caption{Performance of \uid{} for sentiment classification in a federated learning setup.}
    \vspace{-2ex}
    \label{tab:fl}
    \begin{adjustbox}{width=\linewidth, center}
     \newcolumntype{L}{>{\RaggedLeft\arraybackslash}p{0.06\linewidth}} 
  \newcolumntype{O}{>{\RaggedLeft\arraybackslash}m{0.07\linewidth}} 
  \newcolumntype{D}{>{\arraybackslash}m{0.15\linewidth}} 
  \newcolumntype{R}{>{\arraybackslash}m{0.29\linewidth}} 
\begin{tabular}{@{}clSSSSScSSSS@{}}
	\toprule
	& {{Dataset}} & {{Conventional}} 	& {{User Identifier}}		 \\
    \midrule
    \multirow{2}{*}{\STAB{ \scriptsize  RoBERTa}} & 
    IMDB	&44.30 &		47.23 \\
    &Yelp&	68.40 &	70.60			\\
    \midrule[0.1pt]
    \multirow{2}{*}{\STAB{\scriptsize BERT}} &
    Sent140	&84.40	&	86.30		\\
    &Sent140 (Skewed)	 &       86.50	&90.00 	\\

	\bottomrule
\end{tabular}

    \end{adjustbox}
\end{table}

\subsection{\textcolor{black}{Further User-level Accuracy Studies}} \label{sec:change}
\begin{figure}[!htb]
     \centering
     \begin{subfigure}[h]{0.43\textwidth}
         \centering
         \includegraphics[width=\textwidth]{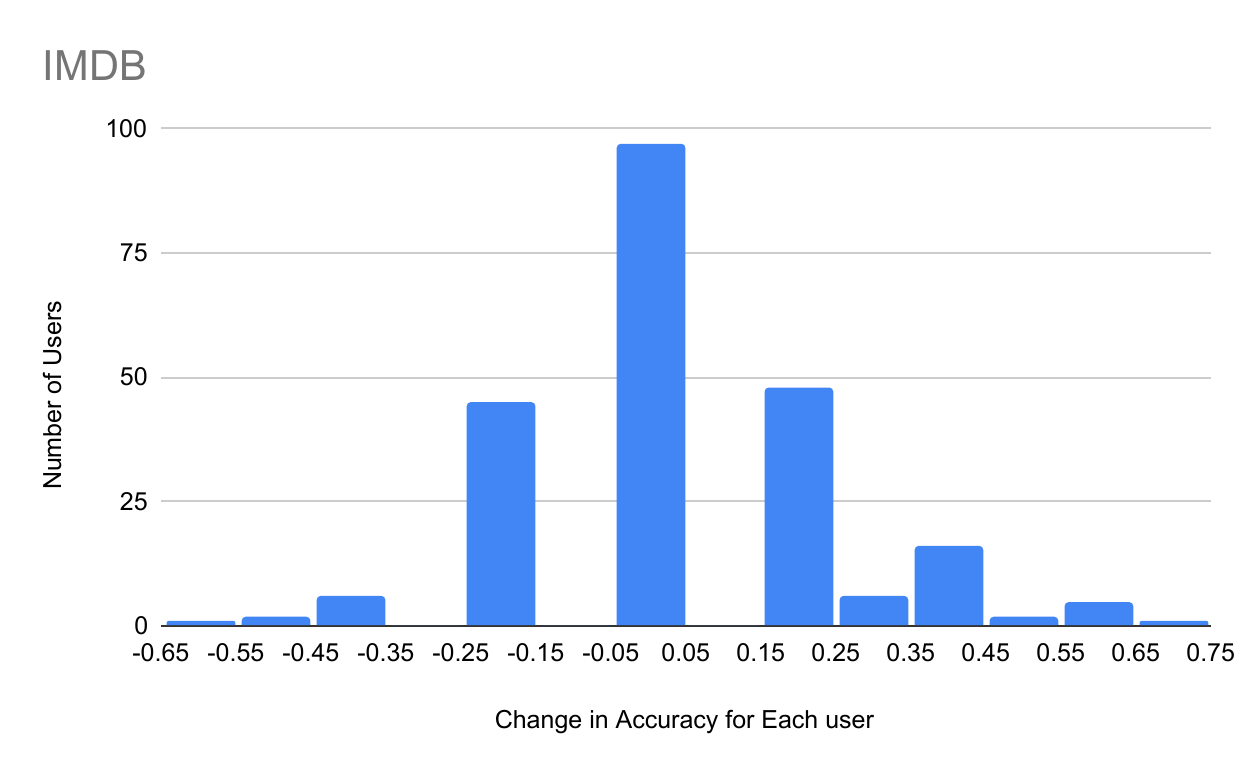}
         \caption{IMDB}
         \label{fig:delta:imdb}
     \end{subfigure}
     ~
     \begin{subfigure}[h]{0.43\textwidth}
         \centering
     \includegraphics[width=\textwidth]{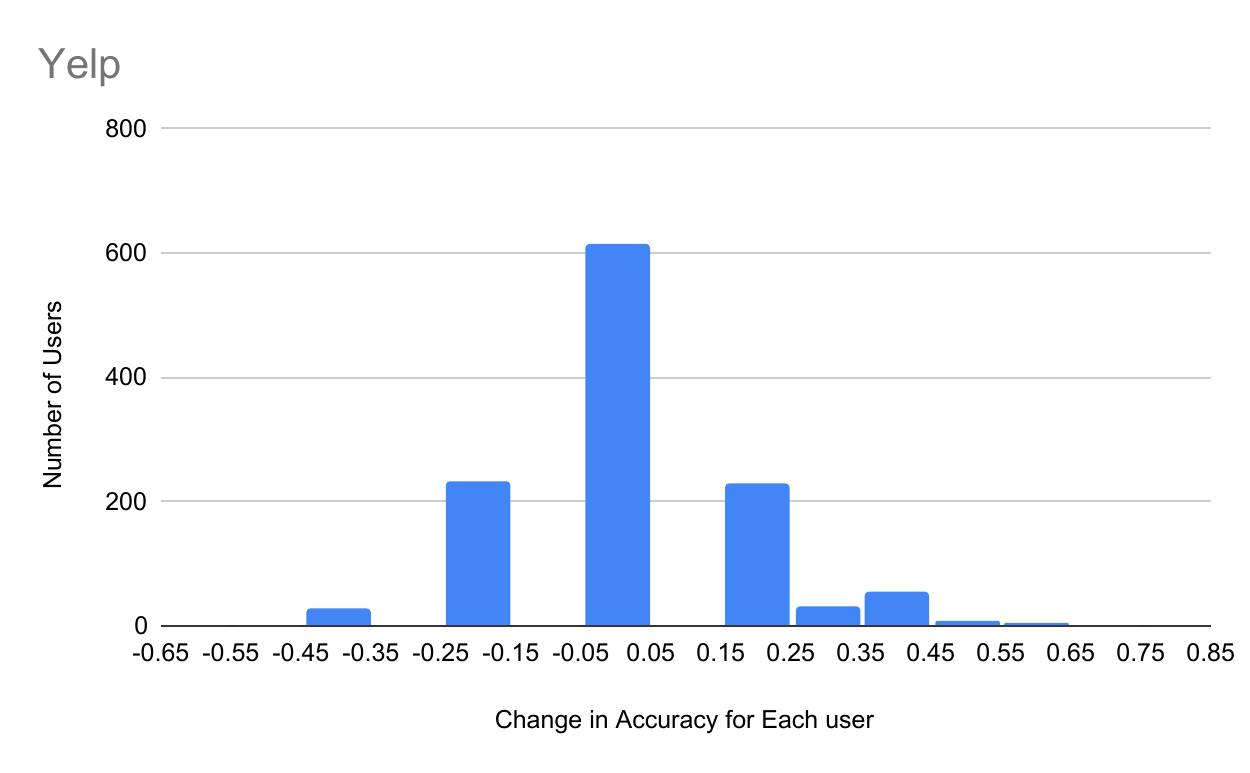}
         \caption{Yelp}
         \label{fig:delta:yelp}
     \end{subfigure}
     \caption{Distribution of test accuracy \textbf{change} across users.
    }
    \vspace{-2ex}
     \label{fig:delta}
\end{figure}

Figure~\ref{fig:delta} shows the change in user accuracy, when we use \uid{} for training, instead of conventional training for each user. In other words, the horizontal axis shows $conventional_{acc}-UID_{acc}$ for each user, and the vertical axis shows the count of users.

As the plots show, on average across the two datasets, $32.1\%$ of the users see improvements in accuracy, whereas $54.2\%$ don't see any change.

\subsection{Maximally Distant User Identifiers}
\textcolor{black}{To better understand the effect of edit distance between user identifiers, We also experimented with \textbf{maximally distanced} identifiers (for the {Rand. All} setup), where the maximum distance would be the length of the identifier here since each token in the identifier can take a substantially large number of values. 
For this experiment, we used rejection sampling for user ids, as in if a new random sample had any token overlaps with existing user ids, we would reject it and sample a new one. 
We observed results very similar to the ones with the random identifiers, which we hypothesize is because the random identifiers are already highly distanced and rarely overlap (less than $10\%$ of the users have non-maximal distance). }

\end{document}